\documentclass{article}

\usepackage[preprint]{neurips_2026}

\usepackage[utf8]{inputenc} 
\usepackage[T1]{fontenc}    
\usepackage{hyperref}       
\usepackage{url}            
\usepackage{booktabs}       
\usepackage{amsfonts}       
\usepackage{nicefrac}       
\usepackage{microtype}      
\usepackage{xcolor}         
\usepackage{wrapfig}
\usepackage{amsmath}
\usepackage{soul}
\sethlcolor{yellow}
\title{Towards Fast and Effective Long Video Understanding of Multimodal Large Language Models via \\Adaptive Quasi-Gaussian Sampling}

\makeatletter
\renewcommand{\@fnsymbol}[1]{%
  \ifcase#1\or \dagger\or \ddagger\or \S\or \P\or \|\or **\or \dagger\dagger\or \ddagger\ddagger \else\@ctrerr\fi}
\makeatother

\author{
\textbf{Kun Zhang} \quad
\textbf{Chenxin Fang} \quad
\textbf{Tao Chen} \quad
\textbf{Baiyang Song} \\
\textbf{Yunhang Shen} \quad
\textbf{Yiyi Zhou}\thanks{Corresponding author.} \quad
\textbf{Rongrong Ji} \\
\\
Key Laboratory of Multimedia Trusted Perception and Efficient Computing, \\
Ministry of Education of China, Xiamen University, 361005, P.R. China \\
\\
\texttt{\{kunzhang,fangchenxin,chentao,songbaiyang,yhshen\}@stu.xmu.edu.cn} \\
\texttt{\{zhouyiyi,rrji\}@xmu.edu.cn}
}
\usepackage{graphicx}
\usepackage{booktabs, multirow, xcolor}
\usepackage{caption}        

\begin{document}

\maketitle

\begin{abstract}

Long video understanding remains a daunting challenge for \emph{Multimodal Large Language Models} (MLLMs) due to the excessive computation and memory footprint. Thus, \emph{keyframe selection} is often adopted to mitigate this shortcoming, which however still suffers from low flexibility and high noise due to its hard sampling principle. In this paper, we define video frame selection as a problem of \emph{Quasi-Gaussian Sampling}, and propose an adaptive and training-free approach termed \textbf{\emph{AdaQ}}. Inspired by the $3$-$\sigma$ rule of Gaussian distribution, the objective of AdaQ is to achieve the optimal $3$-$\sigma$ interval for different examples,  \emph{i.e.}, a smaller $3$-$\sigma$ interval for the local query and a larger one for the global query, thereby facilitating robust and adaptive frame sampling. To validate AdaQ, we apply it to four MLLMs with three embedding models. The extensive experimental results not only show its obvious performance gains over the default MLLMs and the SOTA keyframe selection methods, \emph{e.g.}, helping Qwen3-VL-8B outperform GPT4o by 15.8\% on average by using only 64 frames, but also confirm its superior robustness and high efficiency for long-video understanding, \emph{e.g.}, \textbf{only 1 hyper-parameter} needs to be set. \textbf{Our code project} is given at \href{https://github.com/Zkayovo-xmu/AdaQ}{https://github.com/Zkayovo-xmu/AdaQ}.

\end{abstract}
\section{Introduction}

For a year or two, the research of \emph{Multimodal Large Language Models} (MLLMs) ~\cite{TongLZLS0SJ25,LuoZ0ZSJ25,abs-2411-19628,heurtelcompression,sunds,yanghaplovl,chatgpt4o} has made great breakthroughs in various \emph{vision-language} (VL) tasks. However, long video understanding is still a daunting challenge for existing MLLMs due to the large number of video frames to process, resulting in excessive computation and memory overhead ~\cite{chen2025towards}. For instance, to process 1$k$ frames for an hour-long video, the advanced Qwen3-VL-8B ~\cite{bai2025qwen3vl} requires about 58G GPU memory and 3.28 minutes for one inference. Moreover, the ineffective video understanding not only impedes the performance of MLLMs on video tasks, but also greatly narrows down their application scopes in resource-limited scenarios.  

To this end, \emph{keyframe selection} is arguably a straightforward solution to handle long-video understanding for MLLMs. Inspired by the advancements in LLM-RAG ~\citep{ram2023context,jiang2023active,shi2024replug}, recent endeavors ~\citep{shenlongvu,tang2025adaptive,liu2025boltboostlargevisionlanguage,zhang2025q,chen2025towards,song2026ktv} often adopt keyframe selection strategies to mitigate computational costs by reducing the number of input video frames for MLLMs. In practice, these methods will select a limited number of key frames (or clips) based on the frame-query similarities computed by VL embedding models, \emph{e.g.}, CLIP ~\citep{radford2021learning} or BLIP ~\citep{li2022blip}. 
Consequently, the input sequence length for MLLMs is significantly reduced, while maintaining the ability to perform video understanding tasks effectively using only the selected frames.

Although effective, the existing keyframe selection still suffers from two main limitations. The first one is the flexibility of handling different types of video tasks. Specifically, while keyframe selection is often superior in local video understanding, \emph{e.g.}, \emph{Needle-in-Haystack} ~\cite{abs-2406-08035}, it tends to perform poorly on the global understanding ones. This is because the keyframe selection paradigm often focus on local video snippets, leading to a much narrower vision compared to the default uniform sampling of MLLMs in terms of global understanding. The other issue is the impact of frame-query similarity noise. In the existing frame selection methods ~\cite{tang2025adaptive,zhu2025focus,zhang2025q}, their VL embedding models are commonly pre-trained with plain image-caption pairs, exhibiting obvious gaps to the video examples of MLLMs~\cite{chen2025towards}. Under this rigid selection setting, similarity noise not only declines the precision of frame selections, but also hinders the adaptive adjustment to input queries. Overall, enhancing the flexibility and robustness of frame selection remains a key challenge for MLLMs.

\begin{figure*}[t]
    \centering
    \includegraphics[width=\textwidth]{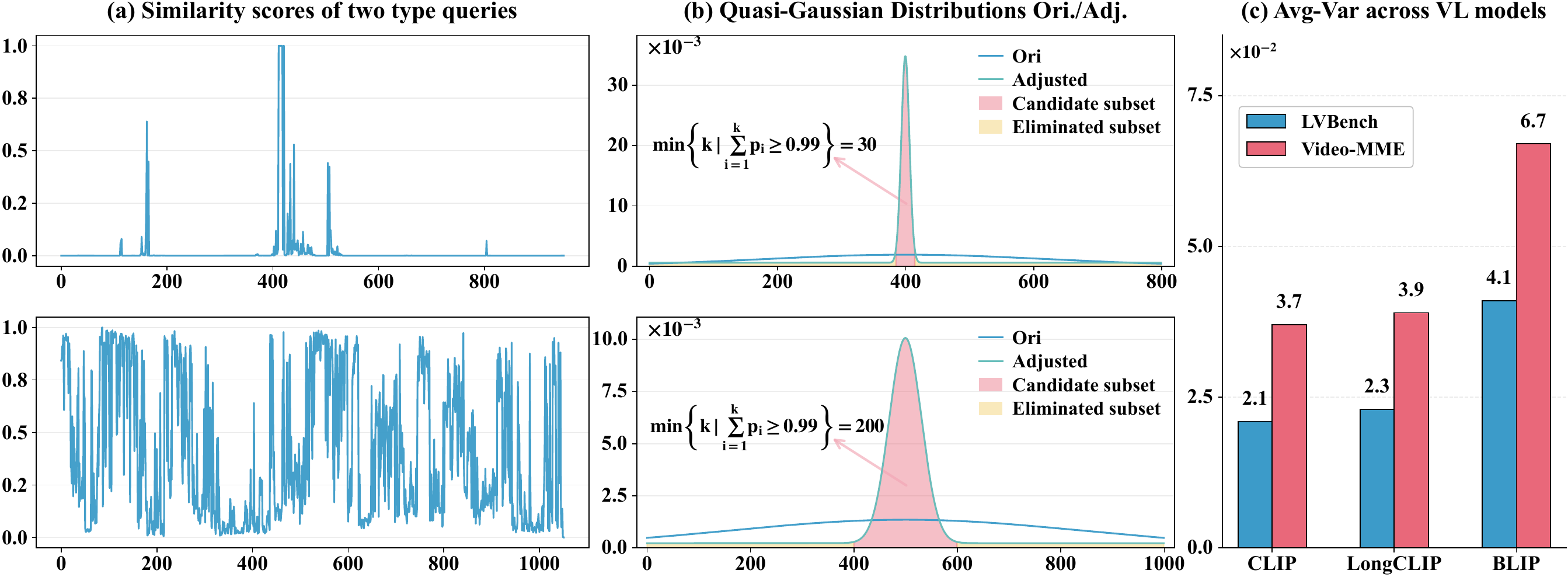}
    \caption{
The statistics of keyframe selection and video examples.
(a) Frame-query similarity scores for two types of queries, \emph{i.e.}, the local (\emph{top}) and global (\emph{bottom}) ones.
(b) The Quasi-Gaussian distributions adjusted by our AdaQ based on the frame-query similarities. Without AdaQ, the default distributions are overly smooth and have excessively large $3$-$\sigma$ intervals.
(c) Average similarity variances of LVBench and VideoMME across different embedding models. Different types of video tasks exhibit obvious distinct frame-query similarity variances. Based on this observation, we use the similarity variance to adaptively adjust the $3$-$\sigma$ intervals for more effective frame sampling.
}
    \label{fig:motivation_v2}
\end{figure*}

To achieve the above target, we redefine video frame selection as a problem of \emph{Adaptive Quasi-Gaussian Sampling}, and propose a novel and training-free method termed \textbf{\emph{AdaQ}} for MLLMs. Specifically, AdaQ is motivated and supported by the classical $3$-$\sigma$ rule of Gaussian distribution, where samples outside the $3$-$\sigma$ interval ($\approx 99.73\%$) are regarded as the low-quality or abnormal ones \cite{casella2002statistical}. To this end, the objective of AdaQ is to build an optimal $3$-$\sigma$ interval for the probabilistic frame sampling, \emph{i.e.}, a larger frame coverage for the global query while a smaller one for the local instruction. However, a naive transformation from query-frame similarities to Gaussian distribution is hard to meet this objective. As shown in Fig. \ref{fig:motivation_v2}, although the similarity scores vary across frames, the large number of frames causes the resulting probabilistic distributions become too flat to distinguish their importances \cite{zhang2025q,TangQXTJY25}.

To address this bottleneck, the design of AdaQ is further supported by an important observation, \emph{i.e.}, the similarity variance of different queries provides a highly informative and robust indicator to distinguish frames. As reported in Fig. \ref{fig:motivation_v2}-c, the average similarity variances of \emph{LVBench} \cite{abs-2406-08035} and \emph{VideoMME} \cite{FuDLLRZWZSZCLLZ25} are significantly different, which are two representative benchmarks for the local and global video understanding, respectively. Motivated by this, we adopt the similarity variance of each example to change the temperature of the \emph{Quasi-Gaussian} distribution, thereby adaptively adjusting $3$-$\sigma$ interval. In this way, AdaQ can adaptively change its sampling strategies to meet the properties of different queries, which also alleviates the noisy impact via its soft sampling manner.

To validate the proposed AdaQ, we apply it to four advanced MLLMs, including LLaVA-OneVision \cite{0080ZGZ00ZZL0L25}, LLaVA-Video-7B \cite{li2024llava}, Qwen2.5-VL \cite{Qwen2.5-VL} and Qwen3-VL \cite{bai2025qwen3vl}, and three VL embedding models, namely CLIP \cite{radford2021learning}, LongCLIP \cite{zhang2024long} and BLIP \cite{li2022blip}. Extensive experiments are conducted on a set of mainstream video benchmarks \cite{WuLCL24,FuDLLRZWZSZCLLZ25,abs-2406-08035,abs-2406-04264}, where a bunch of SOTA keyframe selection methods \cite{liu2025boltboostlargevisionlanguage,tang2025adaptive,zhang2025q} are also comprehensively compared. The experimental results not only witness the consistent advantages of our AdaQ over the default MLLMs as well as the SOTA keyframe selection methods, \emph{e.g.}, improving Qwen3-VL-8B and AKS by +9.0\% and +3.2\% on average, but also confirm its superior robustness and better efficiency for long-video understanding. For instance, AdaQ only requires \textbf{1 hyper-parameter} for most video tasks.

Overall, our contributions are three-fold:
\begin{itemize}
    \item Motivated by the $3$-$\sigma$ rule of Gaussian distribution, we define video frame selection as a problem of Adaptive Quasi-Gaussian Sampling, and propose a novel and training-free framework termed AdaQ for long video understanding of MLLMs.
    \item We identify an important observation that different video tasks exhibit distinct task-wise similarity variance patterns, and further leverage this property to adaptively adjust the Quasi-Gaussian sampling distribution for flexible frame selection.
    \item Extensive experiments on four MLLMs, three VL embedding models, and multiple long-video benchmarks demonstrate that AdaQ consistently outperforms existing keyframe selection methods, while requiring only one hyper-parameter.
\end{itemize}

\section{Related Work}

Recent years have witnessed the rapid development of \emph{multimodal large language models} (MLLMs) towards strong and generalized \emph{vision-language} (VL) capabilities \cite{hurst2024gpt,team2024gemini,LinYZCNJ024,0001RKK24,0080ZGZ00ZZL0L25,li2024llava,Qwen2.5-VL,bai2025qwen3vl,abs-2508-18265}. Despite a big leap forward in image-centric tasks, long video understanding is still challenging for existing MLLMs\cite{liu2024improved}. In particular, existing MLLMs often treat an input video as a set of image patches, each encoded into hundreds of visual tokens \cite{liu2024llavanext}, so the token sequence quickly becomes prohibitively large as videos get longer.


To this end, numerous efforts are recently devoted to efficient long video understanding for MLLMs \cite{kim2025kvzip,alvar2025divprune,kim2024lexico,xu2024think,25commvq,shutova2025cache,feng2024ada,yang2025topv}. One popular solution is to select query-related key frames (clips) for MLLMs \cite{TangQXTJY25,abs-2502-01549,abs-2411-13093,huang2025frag,liu2025boltboostlargevisionlanguage,zhang2025q,chen2025towards,zhu2025focus}. Motivated by LLM-RAG works \cite{shi2024replug,ram2023context,jiang2023active}, keyframe selection methods aim to identify the most informative video frames  based on an external VL embedding model \cite{radford2021learning}, thereby reducing the length of input tokens. There are also some works similar to our probabilistic sampling settings \cite{zhang2025q,liu2025boltboostlargevisionlanguage}. 
In particular, Q-Frame \cite{zhang2025q} also uses \emph{Softmax} to normalize the similarity scores into a probabilistic distribution, based which \emph{Gumbel} noisy is added to enhance the randomness. However,  its sampling is still a hard selection based on the disturbed probabilistic values of frames.   
The principle of BOLT \cite{liu2025boltboostlargevisionlanguage} is more closed to ours, which is also weighted probabilistic sampling. BOLT first adopts the cumulative distribution function to improve the sampling probabilities of key video frames, and then forms a \emph{inverse transform sampling} manner. 
Compared with BOLT, our AdaQ targets at the adaptive and optimal $3$-$\sigma$ interval for frame sampling, which adopts the similarity variances of different examples to  more largely adjust the Quasi-Gaussian distributions.

In addition to the above \emph{frame–query} based methods, another line of research \cite{wang2024videoagent,abs-2411-13093,ma2025drvideo} leverages external tools for MLLMs. There are also alternative methodologies studied for efficient long video understanding of MLLMs, such as \emph{token pruning} \cite{liu2025keyframe,shao2025holitom,yang2025topv,tao2025dycoke} and \emph{KV cache compression} \cite{kim2025kvzip,25commvq,shutova2025cache}. However, our principle and contribution are orthogonal to them, thus they are not in the scope of our comparison.

\section{Preliminary}
\begin{figure*}[t]
  \centering
  \includegraphics[width=\textwidth]{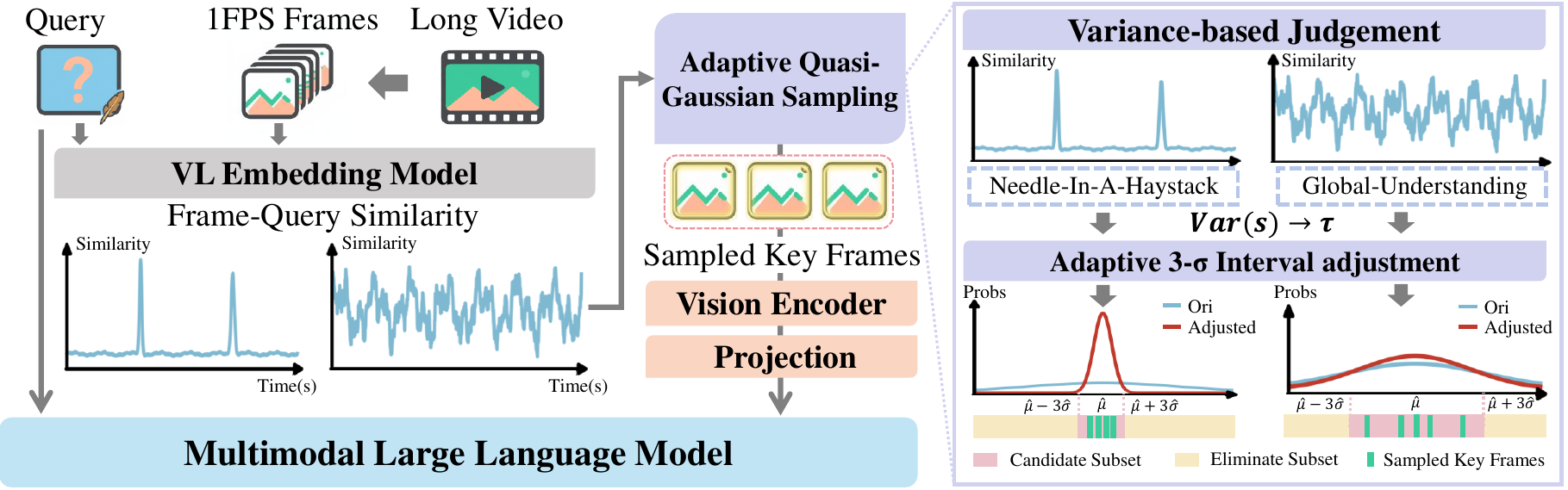}
  \caption{Illustration of the proposed \textbf{AdaQ} approach for the long-video understanding of MLLMs. Given a long video and a user query, a pretrained \emph{Vision-Language} (VL) embedding model is used to compute the frame-query similarities, \emph{e.g.,} CLIP \cite{radford2021learning}. Afterwards, AdaQ will transform these similarity scores into a \emph{Quasi-Gaussian} probability distribution, and adopt the similarity variance, \emph{Var(s)}, to adjust the temperature $\tau$ of the Quasi-Gaussian distribution, thereby achieving adaptive $3$-$\sigma$ interval for different frame samplings.}
  \label{fig:pipline}
\end{figure*}

We first revisit the $3$-$\sigma$ rule of Gaussian distribution. Specifically, given a Gaussian distribution $X \sim \mathcal{N}(\mu, \sigma^2)$, the majority of its probability mass is concentrated within a bounded interval:
\begin{equation}
\mathbb{P}(|X - \mu| \leq 3\sigma) \approx 99.7\%.
\end{equation}
In this case, the effective sampling range is regarded as the $3$-$\sigma$ interval, where samples outside this interval can be regarded as the low-quality or noisy ones ~\cite{casella2002statistical}.

This principle is also applicable to long video understanding. An intuitive solution is to transfer the query-frame similarities of an example into a \emph{Quasi Gaussian} distribution, and define frame selection as a probability sampling task.
\begin{equation}
\label{eq:quasi_gau}
\begin{gathered}
p_i \sim \hat{G}(\hat{\mu},\hat{\sigma}(\tau)), \quad \text{where} \quad
p_i=\frac{\exp\!\left(s_i/\tau\right)}{\sum_{j=1}^T\exp\!\left(s_j/\tau\right)},
\end{gathered}
\end{equation}
where $p_i$ denotes the probability of video frame $V_i$ being sampled, and $s_i$ is the similarity score computed by the VL embedding models like CLIP \cite{radford2021learning}, $T$ denotes the number of candidate frames, and $\hat{G}$ is a discrete \emph{Quasi-Gaussian} distribution. Here, $\tau$ is the temperature factor that controls the shape of the resulting Quasi-Gaussian distribution, the detailed relationship between $\tau$ and the effective sampling range will be further discussed in the next section. In this case, the video frames within the $3$-$\sigma$ interval of $\hat{G}$ can be regarded as the query-relevant ones, while the rest ones will not be considered during sampling. 

However, this direct transformation is hard to meet the target of adaptive and flexible sampling. As shown in Fig. \ref{fig:motivation_v2}, although the similarity scores yield differences across frames, the obtained Quasi Gaussian distribution becomes overly smooth, resulting in an excessively large $3$-$\sigma$ interval. In this case, we focus on achieving the adaptive and optimal $3$-$\sigma$ intervals for different examples, facilitating the flexible and robust frame sampling.
\section{Method}

In this paper, we propose a novel and training-free method termed \emph{AdaQ} towards effective and efficient long video understanding of MLLMs, as illustrated in Fig.~\ref{fig:pipline}.

Given a long video $V$ and a user query $Q$, an MLLM is expected to generate an accurate answer $A$ based on the most informative video frames. A widely used paradigm is keyframe selection based on frame-query similarities:
\begin{equation}
\label{eq:obj}
V_{\text{key}} = \{V_i \mid s_i \in \mathrm{Top}\text{-}K(\{s_t\}_{t=1}^T)\}, \quad s_i = \mathrm{Emb}(V_i, Q).
\end{equation}
Here, $s_i$ denotes the similarity between frame $V_i$ and the query $Q$, and $T$ is the total number of candidate frames. However, this hard selection manner suffers from low flexibility and high noise, especially when the similarity distribution is flat or multi-modal.

To overcome these limitations, we formulate frame selection as an \emph{adaptive Quasi-Gaussian sampling} problem as defined in Eq.\ref{eq:quasi_gau}. Specifically, this Quasi-Gaussian distribution is constructed based on the peak index $\hat{\mu}=\{j\mid p_j=\max P\}$ and the distribution fluctuation coefficient $\hat{\sigma}(\tau)$. In particular, $\hat{\sigma}(\tau)$ controls the effective sampling width of the probability mass over the reordered index set $\mathcal{I}$, and it quantifies how widely the sampling probability mass spreads around the center $\hat{\mu}$.

Inspired by the $3$-$\sigma$ principle as   discussed above, the objective of AdaQ is to construct a bounded high-probability sampling interval around the most relevant region, \emph{i.e.}, $3$-$\sigma$ interval.  Frames inside this interval are preserved as valid candidates, while frames outside it are regarded as low-quality or noisy ones and suppressed from sampling. Therefore, our \emph{Quasi-Gaussian} formulation explicitly models an effective sampling range, which is also the key reason why AdaQ is termed \emph{Quasi-Gaussian Sampling} rather than standard probabilistic sampling.

As shown in Fig.~\ref{fig:motivation_v2}, directly normalizing frame-query similarity scores may lead to an overly flat probability distribution, making it difficult to distinguish truly informative frames from ordinary ones. Under our formulation, a viable solution is to adaptively change the temperature $\tau$ to adjust the distribution curve of $\hat{G}$. More importantly, changing $\tau$ also means adjusting the effective sampling range of the resulting Quasi-Gaussian distribution. But when and how to dynamically set the temperature is still challenging.

To overcome this challenge, we leverage the observed variance pattern of similarity scores to dynamically adjust $\tau$, thereby adaptively reshaping the \emph{Quasi-Gaussian} distribution. Specifically, we define $\tau$ as an adaptive temperature in a variance-driven manner:
\begin{equation}
\label{eq:calc_tau}
\begin{gathered}
\tau=\gamma\cdot\frac{1}{T}\sum_{i=1}^T(\tilde{s}_i-\mu)^2,\quad 
\text{where}\;
\tilde{s}_i=\frac{s_i-\min_j s_j}{\max_k s_k-\min_j s_j},\quad
\mu=\frac{1}{T}\sum_{i=1}^T\tilde{s}_i .
\end{gathered}
\end{equation}

Here, $\gamma$ is a scaling factor for the variance term, and it is used to control the sampling temperature $\tau$, thus adjusting the sharpness of the resulting probability distribution. Notably, $\gamma$ is also the only hyper-parameter of AdaQ. Since $\tau$ determines the spread of the Quasi-Gaussian distribution, this variance-driven design also enables AdaQ to adaptively adjust the corresponding $3$-$\sigma$ range for different queries. In particular, a smaller variance tends to produce a narrower and more concentrated range, while a larger variance leads to a broader range with richer temporal coverage.

Based on this adaptive adjustment, we further introduce a probabilistic sampling strategy to draw keyframes from the adjusted Quasi-Gaussian distribution. Specifically, given the sampling probabilities $P=\{p_i\}_{i\in\mathcal{I}}$, we first determine the effective sampling range over the reordered frame indices according to the bounded high-probability region of $\hat{G}$. Following the intuition of Gaussian sampling, frames outside the $3$-$\sigma$ range are regarded as low-quality candidates and directly suppressed from receiving sampling probability. Then, we perform \emph{probability-weighted sampling} \cite{tille2006sampling}, denoted by $\mathrm{Sample}(\cdot)$, to obtain the final keyframe set $V_{\text{key}}$. Each frame $V_i$ within the effective range is selected with probability proportional to $p_i$, and we sample $K$ keyframes without replacement to encourage both relevance and temporal diversity:
\begin{equation}
\label{eq:sampling}
V_{\text{key}} \sim \mathrm{Sample}\left(\{V_i\}_{i\in Z},\, P_Z,\, K\right),
\end{equation} 
where $Z\subseteq\mathcal{I}$ denotes the valid frame index set within the effective range, and $P_Z$ is the renormalized probability distribution over $Z$. Finally, the sampled frames are sorted by their original timestamps before being fed into the MLLM.

\noindent\textbf{Discussion.} Here, we analyze how the adaptive temperature $\tau$ adjusts the $3$-$\sigma$ interval of the Quasi-Gaussian distribution to achieve proper frame coverage for different types of user queries.

First of all, the temperature $\tau$ adjusts the peak value $p_{\hat{\mu}}$ of this distribution. Specifically, for any $i \neq \hat{\mu}$, we have
\begin{equation}
\frac{p_i}{p_{\hat{\mu}}}
=\exp\!\left(-\frac{s_{\hat{\mu}}-s_i}{\tau}\right),
\qquad \tau \in (0,\infty).
\label{eq:mu}
\end{equation}

A smaller $\tau$ leads to faster exponential decay of non-peak probabilities with a higher peak value, producing a sharper $3$-$\sigma$ interval. In contrast, a larger $\tau$ results in a more uniform distribution with slower decay and a lower peak value, which can be defined by
\begin{equation}
\label{eq:extrme_case}
\lim_{\tau\to+\infty}p_i=\frac{1}{T}, \:\:
\lim_{\tau\to0^+}p_j=
\begin{cases}
1, & j=\hat{\mu}, \\
0, & j\neq \hat{\mu}.  
\end{cases}
\end{equation}

Based on the \emph{cumulative distribution function} (CDF) of the standard Gaussian distribution and Eq.~\ref{eq:mu}, we obtain the following formulation:
\begin{equation}\begin{aligned}P(\hat{\mu}-3\hat{\sigma} \le i \le \hat{\mu}+3\hat{\sigma})= \sum_{i \in Z} p_i= p_{\hat{\mu}} \sum_{i \in Z} \exp\left(-\frac{s_{\hat{\mu}}-s_i}{\tau}\right)= \theta ,\end{aligned}\label{eq:norm_dis}\end{equation}
where $\theta \in [0,1]$ is a probability value, and $Z \subseteq \mathcal{I}$ denotes the valid subset of reordered frame indices within the effective sampling range.

Eq.~\ref{eq:norm_dis} indicates that AdaQ does not simply sample from the full probability space, but instead constructs a bounded high-quality region centered at $\hat{\mu}$, following the $3$-$\sigma$ principle. Therefore, the size of $Z$ directly reflects the temporal coverage of valid candidate frames.

As demonstrated in Eq.~\ref{eq:extrme_case}, we have $\tau \propto -p_{\hat{\mu}}$. Finally, we can obtain the relation between the temperature $\tau$ and the temporal coverage of our proposed Quasi-Gaussian sampling technique according to Eq.~\ref{eq:norm_dis}:
\begin{equation}
|Z|
\propto -\,p_{\hat{\mu}} \exp\!\left(-\frac{s_{\hat{\mu}}-s_i}{\tau}\right)
\propto \tau .
\label{eq:norm_dis_}
\end{equation}

Here, $|Z|$ is the size of the valid frame index set, indicating the coverage of candidate frames, and $\propto$ denotes the \emph{sign} of correlation rather than strict proportionality: $x \propto y$ indicates a positive correlation between $x$ and $y$, whereas $x \propto -y$ indicates a negative correlation.

According to Eq.~\ref{eq:calc_tau} and Eq.~\ref{eq:norm_dis_}, \emph{needle-in-a-haystack} queries typically yield smaller similarity variance and thus smaller $\tau$, resulting in a more compact $|Z|$ and a narrower visual receptive scope. In contrast, global understanding queries tend to have larger variance and larger $\tau$, leading to a larger $|Z|$ and broader temporal coverage. In this case, AdaQ enables flexible and robust frame selection for video MLLMs, making it well suited to real-world scenarios with diverse user queries.
\section{Experiments}
\subsection{Experimental Settings}
\begin{table}[!t]
\centering
\caption{
Comparison between our AdaQ and existing methods on four MLLMs. Our AdaQ consistently outperforms the compared methods under most settings. 
\emph{Frame} denotes the number of sampled video frames. \emph{Avg} is the average gains compared to the default MLLM.  
$^{\dagger}$denotes the reproduced results based on their official codes. The best results are in bold while the second ones are underlined.}
\label{tab:main_results}

\resizebox{\textwidth}{!}{%
\scriptsize
\setlength{\tabcolsep}{2.6pt}
\renewcommand{\arraystretch}{0.95}
\begin{tabular*}{\textwidth}{@{\extracolsep{\fill}} l@{\hspace{0pt}} c@{\hspace{-0.4pt}} c@{\hspace{-0.8pt}} c c c c c c c @{}}
\toprule
\multirow{2}{*}{\textbf{Method}}
& \multirow{2}{*}{\shortstack{\textbf{Embedding}\\\textbf{Model}}}
& \multirow{2}{*}{\textbf{Frames}}
& \multicolumn{2}{c}{\textbf{LongVideoBench}}
& \multicolumn{2}{c}{\textbf{Video-MME}}
& \multirow{2}{*}{\textbf{LVBench}}
& \multirow{2}{*}{\textbf{MLVU}}
& \multirow{2}{*}{\textbf{Avg}}\\
\cmidrule(lr){4-5}
\cmidrule(lr){6-7}
& & & \textbf{Long} & \textbf{Overall} &
 \textbf{Long} & \textbf{Overall} & & & 
\\
\midrule

\multicolumn{1}{l}{\textbf{LLaVA-OneVision-7B}} & - & 32 
& 49.5 & 56.8 
& 48.1 & 58.3 
& 38.6 
& 63.7
& -
\\
\midrule

\textit{Top-k} & CLIP & 32 
& 50.7\,\textcolor{black!45}{(+1.2)} & 58.3\,\textcolor{black!45}{(+1.5)}
& 48.2\,\textcolor{black!45}{(+0.1)} & 58.9\,\textcolor{black!45}{(+0.6)}
& 43.8\,\textcolor{black!45}{(+5.2)}
& 66.9\,\textcolor{black!45}{(+3.2)}
& +5.5\%
\\


\textit{FRAG} & MLLM (7B) & 32 
& - & 57.6\,\textcolor{black!45}{(+0.8)}
& - & 58.6\,\textcolor{black!45}{(+0.3)}
& -
& 65.3\,\textcolor{black!45}{(+1.6)}
& +1.5\%
\\

\textit{OneClip-RAG$^{\dagger}$} & CLIP & 32 
& \underline{54.4}\,\textcolor{black!45}{(+4.9)} & 58.1\,\textcolor{black!45}{(+1.3)}
& 47.7\,\textcolor{black!45}{(-0.4)} & 57.9\,\textcolor{black!45}{(-0.4)}
& 44.1\,\textcolor{black!45}{(+5.5)}
& 67.6\,\textcolor{black!45}{(+3.9)}
& +5.5\%
\\

\textit{AKS} & BLIP & 32 
& 53.0\,\textcolor{black!45}{(+3.5)} & 59.3\,\textcolor{black!45}{(+2.6)}
& \underline{49.3}\,\textcolor{black!45}{(+1.2)} & 58.4\,\textcolor{black!45}{(+0.1)}
& 43.5\,\textcolor{black!45}{(+4.9)}
& 66.8\,\textcolor{black!45}{(+3.1)}
& +5.5\%
\\

\textit{Q-Frame$^{\dagger}_{}$} & LongCLIP & 32 
& 54.0\,\textcolor{black!45}{(+4.5)} & 59.4\,\textcolor{black!45}{(+2.6)}
& 48.3\,\textcolor{black!45}{(+0.2)} & 58.4\,\textcolor{black!45}{(+0.1)}
& \underline{44.3}\,\textcolor{black!45}{(+5.7)}
& \underline{68.5}\,\textcolor{black!45}{(+4.8)}
& +6.8\%
\\

\textit{BOLT} & CLIP & 32 
& - & 59.6\,\textcolor{black!45}{(+2.8)}
& \textbf{49.6}\,\textcolor{black!45}{(+1.5)} & \textbf{59.9}\,\textcolor{black!45}{(+1.6)}
& -
& 66.8\,\textcolor{black!45}{(+3.1)}
& +4.2\%
\\

\midrule

\textit{AdaQ (ours)} & CLIP & 32
& 53.4\,\textcolor{black!45}{(+3.9)} & \underline{59.8}\,\textcolor{black!45}{(+3.0)}
& \textbf{49.6}\,\textcolor{black!45}{(+1.5)} & 59.3\,\textcolor{black!45}{(+1.0)}
& \underline{44.3}\,\textcolor{black!45}{(+5.7)}
& \underline{68.5}\,\textcolor{black!45}{(+4.8)}
& \underline{+7.3\%}
\\

\textit{AdaQ (ours)} & LongCLIP & 32
& \textbf{55.3}\,\textcolor{black!45}{(+5.8)} & \textbf{60.0}\,\textcolor{black!45}{(+3.2)}
& 49.0\,\textcolor{black!45}{(+0.9)} & \underline{59.5}\,\textcolor{black!45}{(+1.2)}
& \textbf{44.8}\,\textcolor{black!45}{(+6.2)}
& \textbf{69.4}\,\textcolor{black!45}{(+5.7)}
& \textbf{+8.2\%}
\\
\midrule

\multicolumn{1}{l}{\textbf{LLaVA-Video-7B}} & - & 64 
& 49.6 & 58.9 
& 53.2 & 64.2 
& 41.9 & 69.5
& -
\\
\midrule

\textit{Top-k} & CLIP & 64 
& 54.6\,\textcolor{black!45}{(+5.0)} & 59.5\,\textcolor{black!45}{(+0.6)}
& 53.2\,\textcolor{black!45}{(+0.0)} & 64.4\,\textcolor{black!45}{(+0.2)}
& 46.9\,\textcolor{black!45}{(+5.0)}
& 71.2\,\textcolor{black!45}{(+1.7)}
& +3.9\%
\\


\textit{FRAG} & MLLM (7B) & 64 
& - & 60.6\,\textcolor{black!45}{(+1.7)}
& - & 63.7\,\textcolor{black!45}{(-0.5)}
& -
& 69.2\,\textcolor{black!45}{(-0.3)}
& +0.6\%
\\

\textit{BOLT} & CLIP & 64 
& - & 62.2\,\textcolor{black!45}{(+3.3)}
& - & 64.6\,\textcolor{black!45}{(+0.4)}
& -
& 70.3\,\textcolor{black!45}{(+0.8)}
& +2.5\%
\\

\textit{E-VRAG} & VLM (2B) & 64 
& - & \underline{63.1}\,\textcolor{black!45}{(+4.2)}
& - & \underline{65.4}\,\textcolor{black!45}{(+1.2)}
& -
& 70.2\,\textcolor{black!45}{(+0.7)}
& +3.3\%
\\

\textit{OneClip-RAG} & CLIP & 64 
& 56.2\,\textcolor{black!45}{(+6.6)} & 62.5\,\textcolor{black!45}{(+3.6)}
& \underline{54.0}\,\textcolor{black!45}{(+0.8)} & 65.2\,\textcolor{black!45}{(+1.0)}
& 48.2\,\textcolor{black!45}{(+6.3)}
& 71.2\,\textcolor{black!45}{(+1.7)}
& +6.3\%
\\

\textit{AKS} & BLIP & 64 
& 54.7\,\textcolor{black!45}{(+5.1)} & 62.7\,\textcolor{black!45}{(+3.8)}
& \textbf{55.0}\,\textcolor{black!45}{(+1.8)} & 65.3\,\textcolor{black!45}{(+1.1)}
& 47.6\,\textcolor{black!45}{(+5.7)}
& 71.8\,\textcolor{black!45}{(+2.3)}
& +6.3\%
\\

\textit{Q-Frame$^{\dagger}_{}$} & LongCLIP & 64 
& 56.9\,\textcolor{black!45}{(+7.3)} & 61.5\,\textcolor{black!45}{(+2.6)}
& 53.9\,\textcolor{black!45}{(+0.7)} & 64.7\,\textcolor{black!45}{(+0.5)}
& 47.1\,\textcolor{black!45}{(+5.2)}
& 72.4\,\textcolor{black!45}{(+2.9)}
& +5.4\%
\\

\midrule

\textit{AdaQ (ours)} & CLIP & 64
& \underline{57.2}\,\textcolor{black!45}{(+7.6)} & \underline{63.1}\,\textcolor{black!45}{(+4.2)}
& \textbf{55.0}\,\textcolor{black!45}{(+1.8)} & \textbf{66.0}\,\textcolor{black!45}{(+1.8)}
& \underline{48.5}\,\textcolor{black!45}{(+6.6)}
& \underline{72.6}\,\textcolor{black!45}{(+3.1)}
& \underline{+7.5\%}
\\
\textit{AdaQ (ours)} & LongCLIP & 64
& \textbf{57.4}\,\textcolor{black!45}{(+8.0)} & \textbf{63.2}\,\textcolor{black!45}{(+4.3)}
& 54.4\,\textcolor{black!45}{(+1.2)} & 65.3\,\textcolor{black!45}{(+1.1)}
& \textbf{49.1}\,\textcolor{black!45}{(+7.2)}
& \textbf{74.2}\,\textcolor{black!45}{(+4.7)}
& \textbf{+8.2\%}
\\
\midrule

\multicolumn{1}{l}{\textbf{Qwen2.5-VL-7B}} & - & 64
& 50.7 & 60.1 
& 52.9 & 63.7 
& 39.3 & 65.5
& -
\\
\midrule

\textit{Top-k} & CLIP & 64 
& 56.4\,\textcolor{black!45}{(+5.7)} & 63.5\,\textcolor{black!45}{(+3.4)}
& 54.8\,\textcolor{black!45}{(+1.9)} & 65.0\,\textcolor{black!45}{(+1.3)}
& 46.7\,\textcolor{black!45}{(+7.4)}
& 70.7\,\textcolor{black!45}{(+5.2)}
& +8.6\%
\\


\textit{OneClip-RAG$^{\dagger}$} & CLIP & 64 
& 55.7\,\textcolor{black!45}{(+5.0)} & 62.2\,\textcolor{black!45}{(+2.1)}
& \textbf{56.3}\,\textcolor{black!45}{(+3.4)} & 65.3\,\textcolor{black!45}{(+1.6)}
& 46.5\,\textcolor{black!45}{(+7.2)}
& 68.6\,\textcolor{black!45}{(+3.1)}
& +7.3\%
\\

\textit{AKS$^\dagger$} & BLIP & 64 
& 56.6\,\textcolor{black!45}{(+5.9)} & 63.8\,\textcolor{black!45}{(+3.7)}
& 54.4\,\textcolor{black!45}{(+1.5)} & 64.6\,\textcolor{black!45}{(+0.9)}
& 46.4\,\textcolor{black!45}{(+7.1)}
& 69.6\,\textcolor{black!45}{(+4.1)}
& +8.0\%
\\

\textit{Q-Frame$^{\dagger}_{}$} & LongCLIP & 64 
& \underline{58.4}\,\textcolor{black!45}{(+7.7)} & \underline{64.8}\,\textcolor{black!45}{(+4.7)}
& 54.0\,\textcolor{black!45}{(+1.1)} & 64.5\,\textcolor{black!45}{(+0.8)}
& 46.5\,\textcolor{black!45}{(+7.2)}
& \underline{72.8}\,\textcolor{black!45}{(+7.3)}
& +9.6\%
\\

\midrule

\textit{AdaQ (ours)} & CLIP & 64
& 57.7\,\textcolor{black!45}{(+7.0)} & 64.4\,\textcolor{black!45}{(+4.3)}
& 55.4\,\textcolor{black!45}{(+2.5)} & \underline{65.5}\,\textcolor{black!45}{(+1.8)}
& \underline{47.0}\,\textcolor{black!45}{(+7.7)}
& 71.8\,\textcolor{black!45}{(+6.3)}
& \underline{+9.8\%}
\\

\textit{AdaQ (ours)} & LongCLIP & 64
& \textbf{58.9}\,\textcolor{black!45}{(+8.2)} & \textbf{65.5}\,\textcolor{black!45}{(+5.4)}
& \underline{55.6}\,\textcolor{black!45}{(+2.7)} & \textbf{65.9}\,\textcolor{black!45}{(+2.2)}
& \textbf{47.1}\,\textcolor{black!45}{(+7.8)}
& \textbf{73.1}\,\textcolor{black!45}{(+7.6)}
& \textbf{+11.0\%}
\\
\midrule

\multicolumn{1}{l}{\textbf{Qwen3-VL-8B}} & - & 64 
& 50.7 & 62.2 
& 56.9 & 67.6 
& 43.6 & 71.0
& -
\\
\midrule

\textit{Top-k} & CLIP & 64 
& 57.6\,\textcolor{black!45}{(+6.9)} & 64.1\,\textcolor{black!45}{(+1.9)}
& 57.3\,\textcolor{black!45}{(+0.4)} & 67.7\,\textcolor{black!45}{(+0.1)}
& 49.2\,\textcolor{black!45}{(+5.6)}
& 73.2\,\textcolor{black!45}{(+2.2)}
& +4.8\%
\\


\textit{OneClip-RAG$^{\dagger}$} & CLIP & 64 
& 58.3\,\textcolor{black!45}{(+7.6)} & 65.1\,\textcolor{black!45}{(+2.9)}
& 56.8\,\textcolor{black!45}{(-0.1)} & 67.3\,\textcolor{black!45}{(-0.3)}
& 50.5\,\textcolor{black!45}{(+6.9)}
& 71.5\,\textcolor{black!45}{(+0.5)}
& +5.2\%
\\

\textit{AKS$^\dagger$} & BLIP & 64 
& 56.9\,\textcolor{black!45}{(+8.0)} & 65.2\,\textcolor{black!45}{(+2.4)}
& 59.0\,\textcolor{black!45}{(+2.1)} & 68.6\,\textcolor{black!45}{(+1.0)}
& 49.0\,\textcolor{black!45}{(+5.6)}
& 74.2\,\textcolor{black!45}{(+3.2)}
& +5.8\%
\\

\textit{Q-Frame$^{\dagger}_{}$} & LongCLIP & 64 
& 58.3\,\textcolor{black!45}{(+7.6)} & 65.0\,\textcolor{black!45}{(+2.8)}
& 57.2\,\textcolor{black!45}{(+0.3)} & 67.9\,\textcolor{black!45}{(+0.3)}
& 50.2\,\textcolor{black!45}{(+6.6)}
& 74.7\,\textcolor{black!45}{(+3.7)}
& +6.3\%
\\

\midrule

\textit{AdaQ (ours)} & CLIP & 64
& \underline{59.1}\,\textcolor{black!45}{(+8.4)} & \underline{66.5}\,\textcolor{black!45}{(+4.3)}
& \textbf{59.6}\,\textcolor{black!45}{(+2.7)} & \textbf{69.8}\,\textcolor{black!45}{(+2.2)}
& \underline{50.6}\,\textcolor{black!45}{(+7.0)}
& \underline{75.0}\,\textcolor{black!45}{(+4.0)}
& \underline{+8.0\%}
\\

\textit{AdaQ (ours)} & LongCLIP & 64
& \textbf{59.8}\,\textcolor{black!45}{(+9.1)} & \textbf{66.9}\,\textcolor{black!45}{(+4.7)}
& \underline{59.4}\,\textcolor{black!45}{(+2.5)} & \underline{69.6}\,\textcolor{black!45}{(+2.0)}
& \textbf{51.4}\,\textcolor{black!45}{(+7.8)}
& \textbf{76.4}\,\textcolor{black!45}{(+5.4)}
& \textbf{+9.0\%}
\\

\bottomrule
\end{tabular*}%
}
\vspace{-0.2cm}
\end{table}
\label{experiments:settings}

\textbf{Benchmarks and metrics.}
We evaluate AdaQ on four mainstream video benchmarks, including
LongVideoBench~\cite{WuLCL24},
Video-MME~\cite{FuDLLRZWZSZCLLZ25},
LVBench~\cite{abs-2406-08035},
and MLVU~\cite{abs-2406-04264}.
LongVideoBench (LVB) and LVBench mainly target at long video understanding with \emph{needle-in-a-haystack} style tasks under extended temporal contexts. And LVBench still has a small proportion of examples about global understanding, as shown in Fig.\ref{fig:ab_qs}. Video-MME emphasizes global understanding and diverse video genres across multiple duration scales.
MLVU is a mixed benchmark that combines both \emph{single-detail} and \emph{holistic} understanding over videos with varied lengths.
We use \emph{Accuracy (Acc)} as the metric.

\textbf{Implementation Details.} 
We validate AdaQ on four advanced MLLMs, including LLaVA-OneVision \cite{0080ZGZ00ZZL0L25}, LLaVA-Video \cite{li2024llava}, Qwen2.5-VL \cite{Qwen2.5-VL} and Qwen3-VL \cite{bai2025qwen3vl}. For each benchmark, we uniformly sample candidate frames at $1$ FPS from the full video, and employ pretrained VL embedding models to compute query--frame similarity scores. The validated embedding models include CLIP \cite{radford2021learning}, LongCLIP \cite{zhang2024long} and BLIP \cite{li2022blip}. Considering the randomness of probabilistic sampling, \textbf{all AdaQ results are averaged over three trials}, and the full statics are provided in the Appendix. Regarding the only hyper-parameter $\gamma$, it is set to 0.5 in most cases, except 1.5 for LongCLIP on VideoMME. As baselines, we adopt uniform sampling (\emph{uniform}) and keyframe selection (\emph{Top-K}), both using a sample frequency of 1 FPS. For compared keyframe selection methods, we use their official codes to reproduce results on new MLLMs and embedding models, \emph{e.g.}, Qwen3-VL and LongCLIP. Unless otherwise specified, all reported AdaQ results are based on the $3$-$\sigma$ interval setting. More analyses about different interval settings are provided in the Appendix~\ref{app:interval}.

\subsection{Experimental Results}
\subsubsection{Quantitative Analysis}
\begin{figure}[t]
    \centering
    \includegraphics[width=\linewidth]{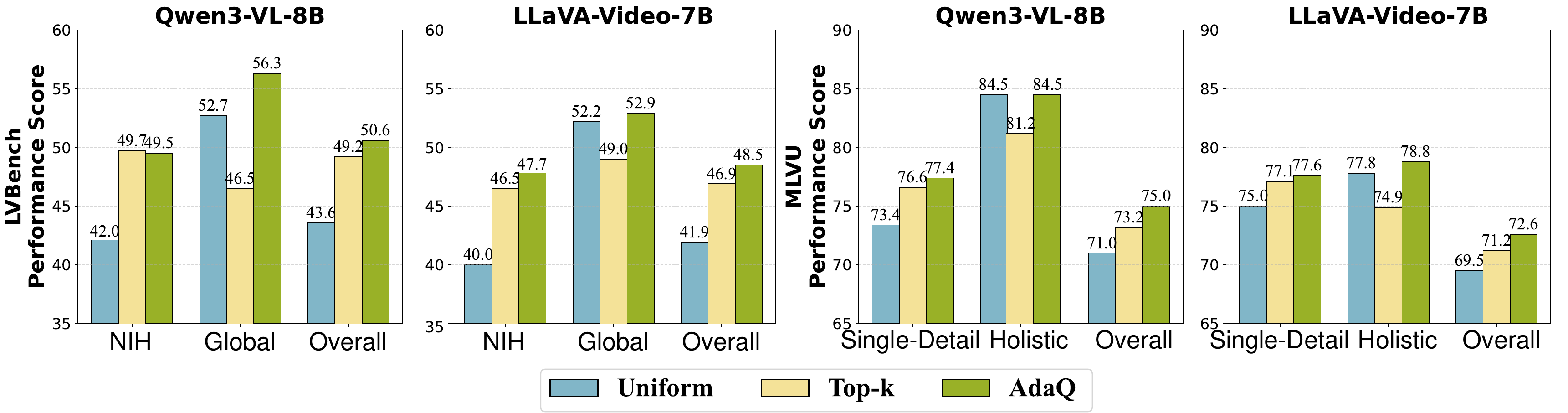}
    \caption{
Comparison between AdaQ, keyframe selection (\emph{Top-$K$})  and uniform sampling. Keyframe selection often perform well on the local understanding tasks, \emph{i.e.}, \emph{needle-in-a-haystack} (NIH) and \emph{single-detail}, but it is inferior in the global ones due to its rigid sampling strategy. In contrast, our AdaQ is capable of adaptive and flexible sampling for different queries.
} \label{fig:ab_qs}
\end{figure}
\begin{figure}[t]
    \centering
    \begin{minipage}[c]{0.5\linewidth}
        \centering
        \captionof{table}{Comparisons with different VL embedding models. \emph{LCLIP} denotes LongCLIP. The number of frames is 64. In comparison, AdaQ is more robust to embedding models.}
        \label{tab:ablation_qwen_clip_blip}
        \scriptsize
        \setlength{\tabcolsep}{2.4pt}
        \renewcommand{\arraystretch}{0.95}
        \resizebox{\linewidth}{!}{%
        \begin{tabular}{lcccccc}
            \toprule
            \multirow{2}{*}{\textbf{Method}} &
            \multicolumn{3}{c}{\textbf{LongVideoBench}} &
            \multicolumn{3}{c}{\textbf{MLVU}} \\
            \cmidrule(lr){2-4}\cmidrule(lr){5-7}
            & \textbf{CLIP} & \textbf{BLIP} & \textbf{LCLIP}
            & \textbf{CLIP} & \textbf{BLIP} & \textbf{LCLIP} \\
            \midrule
            \textbf{LLaVA-Video}
            &  & 58.9 &  &  & 69.5 &  \\
            \midrule
            Top-k
            & 59.5\,\textcolor{black!45}{+0.6}
            & 61.3\,\textcolor{black!45}{+2.4}
            & 62.0\,\textcolor{black!45}{+3.1}
            & 71.0\,\textcolor{black!45}{+1.5}
            & 72.4\,\textcolor{black!45}{+2.9}
            & 73.0\,\textcolor{black!45}{+3.5} \\
            Q-Frame
            & 62.4\,\textcolor{black!45}{+3.5}
            & 62.6\,\textcolor{black!45}{+3.7}
            & 62.1\,\textcolor{black!45}{+3.2}
            & 71.5\,\textcolor{black!45}{+2.0}
            & 72.8\,\textcolor{black!45}{+3.3}
            & 72.9\,\textcolor{black!45}{+3.4} \\
            AKS
            & 62.2\,\textcolor{black!45}{+3.3}
            & 62.7\,\textcolor{black!45}{+3.8}
            & 62.0\,\textcolor{black!45}{+3.1}
            & 71.2\,\textcolor{black!45}{+1.7}
            & 71.8\,\textcolor{black!45}{+2.3}
            & 72.0\,\textcolor{black!45}{+2.5} \\
            AdaQ
            & \textbf{63.1}\,\textcolor{black!45}{+4.2}
            & \textbf{63.2}\,\textcolor{black!45}{+4.3}
            & \textbf{63.2}\,\textcolor{black!45}{+4.3}
            & \textbf{72.0}\,\textcolor{black!45}{+2.5}
            & \textbf{73.4}\,\textcolor{black!45}{+3.9}
            & \textbf{74.2}\,\textcolor{black!45}{+4.7} \\
            \midrule
            \textbf{Qwen3-VL}
            &  & 62.2 &  &  & 71.0 &  \\
            \midrule
            Top-k
            & 64.1\,\textcolor{black!45}{+1.9}
            & 65.1\,\textcolor{black!45}{+2.9}
            & 64.7\,\textcolor{black!45}{+2.5}
            & 73.2\,\textcolor{black!45}{+2.2}
            & 74.2\,\textcolor{black!45}{+3.2}
            & 74.6\,\textcolor{black!45}{+3.6} \\
            Q-Frame
            & 64.7\,\textcolor{black!45}{+2.5}
            & 66.0\,\textcolor{black!45}{+3.8}
            & 65.0\,\textcolor{black!45}{+2.8}
            & 73.9\,\textcolor{black!45}{+2.9}
            & 74.0\,\textcolor{black!45}{+3.0}
            & 74.7\,\textcolor{black!45}{+3.7} \\
            AKS
            & 64.6\,\textcolor{black!45}{+2.4}
            & 65.2\,\textcolor{black!45}{+3.0}
            & 66.0\,\textcolor{black!45}{+3.8}
            & 73.1\,\textcolor{black!45}{+2.1}
            & 73.9\,\textcolor{black!45}{+2.9}
            & 75.3\,\textcolor{black!45}{+4.3} \\
            AdaQ
            & \textbf{66.5}\,\textcolor{black!45}{+4.3}
            & \textbf{66.7}\,\textcolor{black!45}{+4.5}
            & \textbf{66.9}\,\textcolor{black!45}{+4.7}
            & \textbf{75.0}\,\textcolor{black!45}{+4.0}
            & \textbf{75.4}\,\textcolor{black!45}{+4.4}
            & \textbf{76.4}\,\textcolor{black!45}{+5.4} \\
            \bottomrule
        \end{tabular}%
        }
    \end{minipage}
    \hfill
    \begin{minipage}[c]{0.48\linewidth}
        \centering
        \captionof{table}{Ablation of the number of sampled video frames.}
        \label{tab:ablation_frame_}
        \scriptsize
        \setlength{\tabcolsep}{2.0pt}
        \renewcommand{\arraystretch}{0.95}
        \resizebox{\linewidth}{!}{%
        \begin{tabular}{lcccccc}
            \toprule
            \textbf{Model} & \textbf{Frames} & \textbf{LVB} & \textbf{Video-MME} & \textbf{LVBench} & \textbf{MLVU} & \textbf{Avg} \\
            \midrule
            \multicolumn{7}{l}{\textbf{Qwen3-VL-8B}} \\
            \cmidrule(lr){1-7}
            Uniform  & 32  & 59.8 & 64.5 & 40.6 & 67.1 & \textcolor{black!45}{-}\\
            Top-k    & 32  & 62.2\,\textcolor{black!45}{+2.4} & 66.1\,\textcolor{black!45}{+1.6} & 43.4\,\textcolor{black!45}{+2.8} & 70.4\,\textcolor{black!45}{+3.3} & \textcolor{black!45}{+4.6\%}\\
            AdaQ     & 32  & 63.8\,\textcolor{black!45}{+4.0} & 67.7\,\textcolor{black!45}{+3.2} & 46.8\,\textcolor{black!45}{+6.2} & 72.0\,\textcolor{black!45}{+4.9} & \textcolor{black!45}{+8.6\%}\\
            \cmidrule(lr){1-7}
            Uniform  & 64  & 62.2 & 66.9 & 43.6 & 71.0 & \textcolor{black!45}{-}\\
            Top-k    & 64  & 64.1\,\textcolor{black!45}{+1.9} & 68.1\,\textcolor{black!45}{+1.2} & 49.2\,\textcolor{black!45}{+5.6} & 73.2\,\textcolor{black!45}{+2.2} & \textcolor{black!45}{+5.2\%}\\
            AdaQ     & 64  & 66.5\,\textcolor{black!45}{+4.3} & 69.8\,\textcolor{black!45}{+2.9} & 50.6\,\textcolor{black!45}{+7.0} & 75.0\,\textcolor{black!45}{+4.0} & \textcolor{black!45}{+8.2\%}\\
            \cmidrule(lr){1-7}
            Uniform  & 128 & 64.8 & 69.9 & 47.6 & 74.7 & \textcolor{black!45}{-}\\
            Top-k    & 128 & 66.6\,\textcolor{black!45}{+1.8} & 70.2\,\textcolor{black!45}{+0.3} & 51.8\,\textcolor{black!45}{+4.2} & 75.3\,\textcolor{black!45}{+0.6} & \textcolor{black!45}{+3.2\%}\\
            AdaQ     & 128 & 66.8\,\textcolor{black!45}{+2.0} & 70.6\,\textcolor{black!45}{+0.7} & 52.8\,\textcolor{black!45}{+5.2} & 76.4\,\textcolor{black!45}{+1.7} & \textcolor{black!45}{+4.3\%}\\
            \cmidrule(lr){1-7}
            Uniform  & 256 & 65.7 & 70.5 & 52.0 & 77.0 & \textcolor{black!45}{-}\\
            Top-k    & 256 & 66.7\,\textcolor{black!45}{+1.0} & 70.9\,\textcolor{black!45}{+0.4} & 54.5\,\textcolor{black!45}{+2.5} & 77.3\,\textcolor{black!45}{+0.3} & \textcolor{black!45}{+1.8\%}\\
            AdaQ     & 256 & 67.7\,\textcolor{black!45}{+2.0} & 71.3\,\textcolor{black!45}{+0.8} & 55.0\,\textcolor{black!45}{+3.0} & 77.7\,\textcolor{black!45}{+0.7} & \textcolor{black!45}{+2.7\%}\\
            \bottomrule
        \end{tabular}%
        }
    \end{minipage}
\end{figure}

\noindent\textbf{Comparison with existing keyframe selection methods.}
We first compare our AdaQ to existing keyframe selection methods on four MLLMs, as reported in Tab.~\ref{tab:main_results}. We first observe that the naive keyframe selection method is a strong baseline despite its simplicity. But its advantages mainly lie in the local understanding tasks as discussed above, \emph{e.g.}, LongVideo and LVBench. Similar cases can also be seen on recent approach OneClip-RAG, which focuses on selecting relevant clips via frame-query similarities. In comparison, advanced keyframe selection methods often exhibit more balanced performance across tasks. For instance, AKS and Q-Frame obtain decent results on both LVBench and MLVU, mainly due to their adaptive and flexible sampling designs. Compared with these SOTA methods, our AdaQ achieves the best performance under most settings with average improvements of 3.2\% and 2.7\% on Qwen3-VL, respectively. More importantly, our advantages are consistent across MLLMs.
Also shown in Tab. \ref{tab:ablation_qwen_clip_blip}, AdaQ still outperforms the two SOTA methods on all three embedding models. 
Overall, these results greatly validate our superiority and robustness.

\noindent\textbf{Analysis of keyframe selection.}
We further make a thorough comparison to keyframe selection (\emph{Top-$K$}) in Fig.~\ref{fig:ab_qs} and Tab. \ref{tab:ablation_qwen_clip_blip}-\ref{tab:ablation_frame_}. The results of Fig. \ref{fig:ab_qs} confirm our argument about the inferior flexibility of keyframe selection. Although keyframe selection performs well on the local understanding tasks of LVBench and MLVU, \emph{i.e.}, \emph{NIH} and \emph{Single-Detail}, it also suffers from obvious drops in the global (holistic) ones, \emph{e.g.}, -6.2\% than uniform sampling (\emph{Uniform}) of Qwen3-VL-8B on LVBench-Global. In stark contrast, AdaQ obtains consistent gains for both local and global tasks compared to the default uniform sampling, \emph{e.g.}, it improves 3.1\% and 4.0\% under LLaVA-Video-7B and Qwen3-VL-8B on MLVU, respectively. These results well confirm the superior flexibility of our AdaQ for video tasks.  

Tab. \ref{tab:ablation_qwen_clip_blip} illustrates the dependence of keyframe selection on the capabilities of VL embedding models. Given a weaker embedding model like CLIP, LLaVA-Video-7B receives obviously worse results on the video benchmarks, especially in the local understanding task where it excels, \emph{e.g.}, -2.5\% compared to LongCLIP on LongVideoBench. 
These results also confirm our argument about the similarity noise impact of keyframe selection in terms of its hard selection manner. Compared with it, our AdaQ shows much better robustness to embedding models. Tab. \ref{tab:ablation_frame_} shows the results of keyframe selection and our AdaQ using different numbers of sampled frames. It can be seen that  using more video frames is a solution to mitigate the shortcomings of keyframe selection, but its gains to uniform sampling also become marginal. In comparison, our AdaQ reaches the performance upper-bound by using much fewer frames.
Overall, these results not only confirm the argued shortcomings of keyframe selection, but also confirm the soft and adaptive principle of our AdaQ.

\begin{wrapfigure}{t}{0.54\textwidth}
  \vspace{-13pt}
  \centering
  \captionof{table}{
  Comparison between \emph{Qwen3-VL+AdaQ} and existing SOTA MLLMs.
  With AdaQ, Qwen3-VL can outperform several larger MLLMs.
  }
  \label{tab:main}
  \vspace{-6pt}

  \resizebox{\linewidth}{!}{%
    \small
    \setlength{\tabcolsep}{2.6pt}
    \renewcommand{\arraystretch}{0.95}
    \begin{tabular}{l c c c c c}
      \toprule
      \textbf{Method} & \textbf{Frames} & \textbf{LVB} & \textbf{Video-MME} & \textbf{LVBench} & \textbf{MLVU} \\
      \midrule
      GPT-4o & 256/0.5fps & 66.7 & 71.9 & 34.7 & 64.6 \\
      Gemini-1.5-Pro & 256 & 64.0 & 75.0 & 33.1 & - \\
      \midrule
      Qwen2.5-VL-72B & 1fps & 60.7 & \textbf{73.3} & 47.3 & 74.6 \\
      Kimi-VL-16B-A3B & 64 & 64.5 & 67.8 & - & 74.2 \\
      LLaVA-Video-72B & 64 & 63.9 & 70.0 & 45.5 & 74.4 \\
      InternVL3.5-38B & 64 & 65.7 & 70.9 & - & 77.0 \\
      \midrule
      \textbf{Qwen3-VL-8B} & 64 & 62.2 & 66.9 & 43.6 & 71.0 \\
      \textbf{+AdaQ} & 64 & 66.9 & 69.4 & 51.4 & 76.4 \\
      \midrule
      \textbf{Qwen3-VL-30B} & 64 & 67.2 & 69.9 & 44.0 & 72.8 \\
      \textbf{+AdaQ} & 64 & \textbf{70.2} & 71.1 & \textbf{52.4} & \textbf{77.8} \\
      \bottomrule
    \end{tabular}
  }

  \vspace{-15pt}
\end{wrapfigure}
\noindent\textbf{Comparison with SOTA Video-MLLMs.} In Tab.~\ref{tab:main}, we also apply AdaQ to two Qwen3-VL models and compare them with a set of SOTA video-MLLMs.
It can be seen that although Qwen3-VL series are advanced and powerful MLLMs, they still lag behind other MLLMs which uses larger parameter size or more video frames. These two factors are critical for existing MLLMs to pursuit stronger video capabilities. 
However, with the help of AdaQ, Qwen3-VL-8B and Qwen3-VL-30B can boost their performance on all benchmarks, even outperforming the close-source MLLMs by a large margin. For instance, Qwen3-VL-30B obtains 3.5\%, 17.7\% and 13.2\% gains over GPT-4o on LVB, LVBench and MLVU, respectively. 
These notable results further validate the effectiveness of our AdaQ and also solidify our contribution to the community.

\begin{wrapfigure}{r}{0.48\textwidth}
  \vspace{-12pt}
  \centering

  \includegraphics[width=\linewidth]{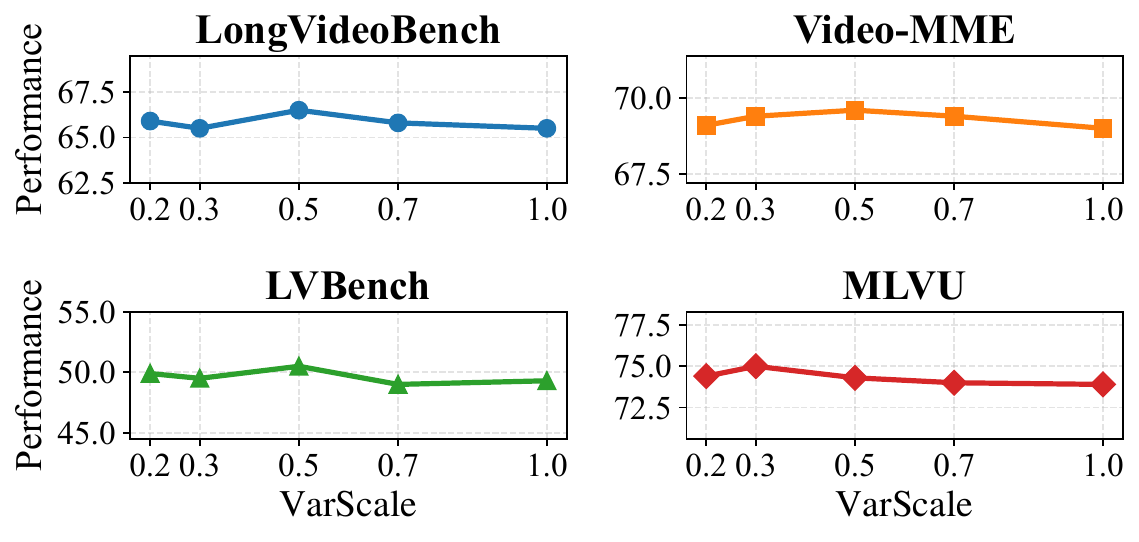}
  \vspace{0pt}
  \caption{
  Ablation of hyper-parameter $\gamma$.
  AdaQ is robust and $\gamma=0.5$ works well across settings.
  }
  \label{fig:ablation_varscale}

  \vspace{-8pt}
\end{wrapfigure}
\noindent\textbf{Ablation Studies of hyper-parameter.}
 In Fig.~\ref{fig:ablation_varscale}, we report the sensitivity of AdaQ to its only hyper-parameter, \emph{i.e.}, $\gamma$ in Eq. \ref{eq:calc_tau} for adjusting the Quasi-Gaussian distribution. 
From these plots, we can see that AdaQ is not very sensitive to the selection of $\gamma$. Although it changes still bring slight  performance variations, a direct choice of 0.5 can already obtain high performance for all MLLMs on most benchmarks. These results demonstrate that our AdaQ is a novel and neat method, and its simple and robust setup can well facilitate its practical use. \textbf{More analyses are in Appendix. }
\begin{figure*}[t]
  \centering
  \includegraphics[width=\textwidth]{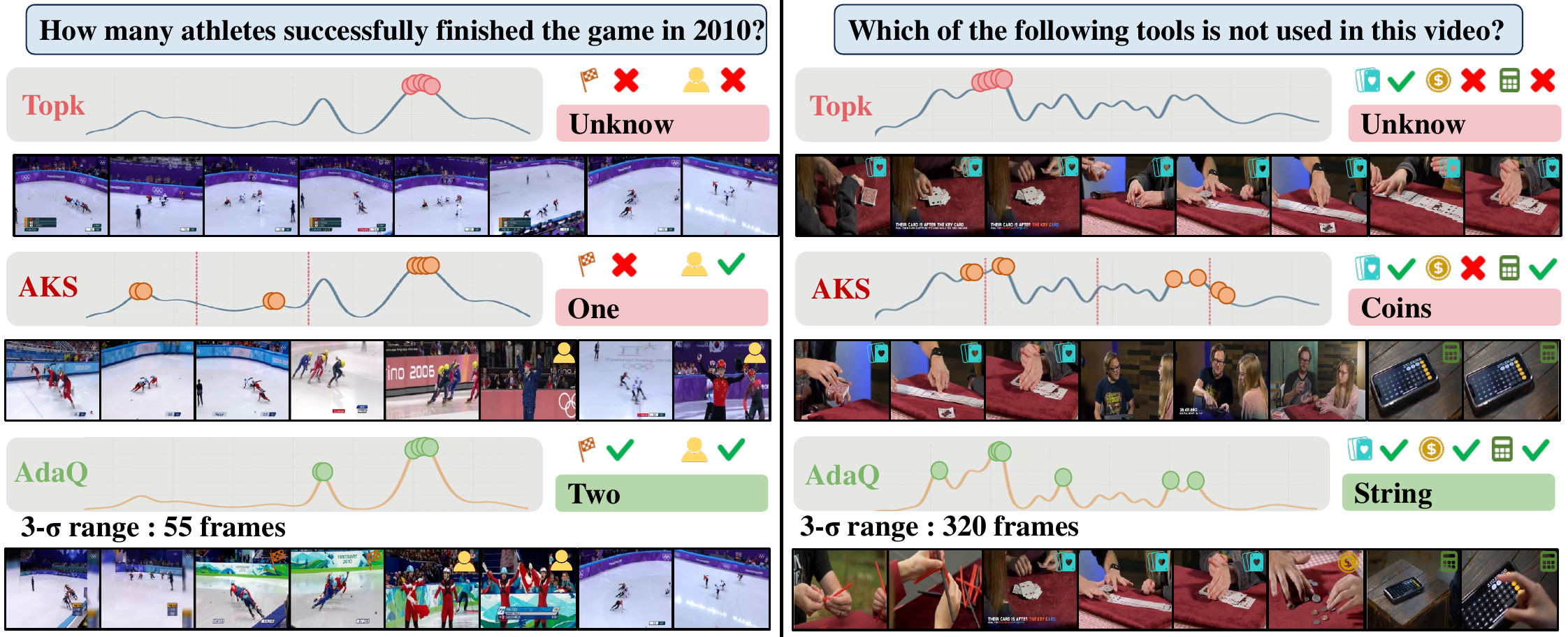}
  \caption{Visualized comparisons between Top-k, AKS and our AdaQ. For each example, their selected frames are given at the bottom, and the curves are the similarity distribution for \emph{Top-$K$} and AKS, while the ones of AdaQ are the adjusted probability distribution. The right pannels indicate that whether the required cues are covered, \emph{e.g.}, calculator and playing cards. In comparison, AdaQ can provide more adaptive sampling strategies to obtain more  suitable temporal coverages.}
  \label{fig:Visualization}
\end{figure*}

\subsubsection{Qualitative Analysis}

In Fig.~\ref{fig:Visualization}, we visualize the sampled frames and prediction of keyframe selection (\emph{Top-$k$}), AKS and AdaQ. For each example, the right panels show whether the required clues are covered by retrieved visual evidence, \emph{e.g.}, calculator and playing cards. As observed, AdaQ consistently retrieves more instruction-relevant frames with better temporal coverage, leading to more complete evidence grounding.
In contrast, Top-$k$ and AKS may miss key clues or over-focus on local peaks, resulting in fragmented evidence and unreliable reasoning.
Conclusively, AdaQ enables the MLLM to answer more accurately in both multi-detail reasoning (left) and global understanding (right) tasks.

\setlength{\intextsep}{0pt}

\section{Conclusion}
In this paper, we propose a novel and training-free approach termed AdaQ for the effective long video understanding of MLLMs. Different from existing hard selection approaches, AdaQ defines video frame selection as a  problem of \emph{Adaptive Quasi-Gaussian sampling}, and applies the observed task-wise similarity variance to adaptively change the optimal $3$-$\sigma$ intervals, thereby achieving a flexible and robust sampling strategies for different queries. The extensive experiments on 4 MLLMs and 3 VL embedding models not only witness its superiority than the default MLLMs and the compared methods, \emph{e.g.}, helping Qwen3-VL-8B outperform GPT4o by 15.8\% on average, but also well confirm its robustness to real-world applications.

\clearpage



\bibliographystyle{plainnat} 
\bibliography{ref}
\clearpage

\appendix
\section{LVBench NIH/Global Subset Analysis}
\begin{table}[!t]
\centering
\caption{
{Comparison of different post-sampling interval settings on \emph{Qwen3-VL-8B} across four long-video benchmarks.
\emph{Probs-Ori} samples from the original probability distribution over all frames, while \emph{Probs-Adjusted} uses the variance-adjusted distribution.
\emph{AdaQ} further restricts sampling to a high-probability interval, where $2$-$\sigma$ and $3$-$\sigma$ denote intervals with different cumulative probability coverage.
\emph{Avg} denotes the average relative gain over uniform sampling.
The best results are in bold.}}
\label{tab:interval_results}

\resizebox{\textwidth}{!}{%
\scriptsize
\setlength{\tabcolsep}{2.6pt}
\renewcommand{\arraystretch}{0.95}
\begin{tabular*}{\textwidth}{@{\extracolsep{\fill}} l c c c c c c @{}}
\toprule
\textbf{Method}
& \textbf{Interval}
& \textbf{LongVideoBench}
& \textbf{Video-MME}
& \textbf{LVBench}
& \textbf{MLVU}
& \textbf{Avg}
\\
\midrule

\textit{Uniform} & - 
& 62.2
& 67.6
& 43.6
& 71.0
& -
\\

\textit{Probs-Ori} & All 
& 62.5\,\textcolor{black!45}{(+0.3)}
& 68.0\,\textcolor{black!45}{(+0.4)}
& 44.1\,\textcolor{black!45}{(+0.5)}
& 70.8\,\textcolor{black!45}{(-0.2)}
& +0.5\%
\\

\textit{Probs-Adjusted} & All 
& 66.0\,\textcolor{black!45}{(+3.8)}
& 69.4\,\textcolor{black!45}{(+1.8)}
& 50.0\,\textcolor{black!45}{(+6.4)}
& 74.8\,\textcolor{black!45}{(+3.8)}
& +7.2\%
\\

\textit{AdaQ} & $2$-$\sigma$
& 65.7\,\textcolor{black!45}{(+3.5)}
& 69.1\,\textcolor{black!45}{(+1.5)}
& \textbf{51.9}\,\textcolor{black!45}{(+8.3)}
& 74.6\,\textcolor{black!45}{(+3.6)}
& \textbf{+8.0\%}
\\

\textit{AdaQ} & $3$-$\sigma$
& \textbf{66.5}\,\textcolor{black!45}{(+4.3)}
& \textbf{69.8}\,\textcolor{black!45}{(+2.2)}
& 50.6\,\textcolor{black!45}{(+7.0)}
& \textbf{75.0}\,\textcolor{black!45}{(+4.0)}
& \textbf{+8.0\%}
\\

\bottomrule
\end{tabular*}%
}
\vspace{-0.2cm}
\end{table}
To better interpret the trends in Fig.~\ref{fig:ab_qs}, we additionally report results on two LVBench subsets split by the official \emph{time\_reference} annotation, which indicates the temporal span of the ground-truth evidence. Specifically, we categorize instances with \emph{time\_reference} $>4$ minutes as \emph{global understanding} (Global), where answering typically requires aggregating information across a long temporal context. The remaining instances are treated as \emph{Needle-In-A-Haystack} (NIH), in which the required clue lies in a relatively short window and thus emphasizes precise evidence localization. This subset analysis helps reveal whether the performance changes in Fig.~\ref{fig:ab_qs} stem from improved long-range coverage (Global) or more accurate retrieval of sparse yet critical clues (NIH).

\section{Additional Experimental Details}

\subsection{Comparison of different post-sampling interval settings}
\label{app:interval}
In Tab.~\ref{tab:interval_results}, we further study the effect of different post-sampling interval settings. Directly sampling from the original probability distribution over all frames, i.e., \emph{Probs-Ori}, brings only marginal gains over uniform sampling, indicating that the original similarity-induced distribution is still too flat to provide effective frame selection. After applying our variance-based adjustment, \emph{Probs-Adjusted} achieves much stronger improvements, e.g., +7.2\% average gain, which verifies the importance of adaptively reshaping the Quasi-Gaussian distribution. Moreover, by further restricting sampling to a high-probability interval, AdaQ achieves consistently strong performance under both $2$-$\sigma$ and $3$-$\sigma$ settings. In particular, the $3$-$\sigma$ interval achieves the best results on LongVideoBench, Video-MME, and MLVU, while the $2$-$\sigma$ interval performs better on LVBench. These results demonstrate that suppressing low-probability frames outside the effective interval is beneficial, and the broader $3$-$\sigma$ setting provides a robust default choice across different long-video benchmarks.

\begin{table*}[t]
  \centering
  \caption{Per-run results (test1/2/3) and mean scores across benchmarks in Tab. \ref{tab:main_results}.}
  \label{tab:flow}

  \tiny
  \setlength{\tabcolsep}{2.0pt}
  \renewcommand{\arraystretch}{0.9}
  \begin{tabular*}{\textwidth}{@{\extracolsep{\fill}} l c cccc cccc cccc cccc @{}}
    \toprule
    \multirow{2}{*}{\textbf{Model}}
    & \multirow{2}{*}{\shortstack{\textbf{Embedding}\\\textbf{Model}}}
    & \multicolumn{4}{c}{\textbf{LongVideoBench}}
    & \multicolumn{4}{c}{\textbf{Video-MME}}
    & \multicolumn{4}{c}{\textbf{LVBench}}
    & \multicolumn{4}{c}{\textbf{MLVU}} \\
    \cmidrule(lr){3-6} \cmidrule(lr){7-10} \cmidrule(lr){11-14} \cmidrule(lr){15-18}
    &
    & \textbf{t1} & \textbf{t2} & \textbf{t3} & \textbf{mean}
    & \textbf{t1} & \textbf{t2} & \textbf{t3} & \textbf{mean}
    & \textbf{t1} & \textbf{t2} & \textbf{t3} & \textbf{mean}
    & \textbf{t1} & \textbf{t2} & \textbf{t3} & \textbf{mean} \\
    \midrule

    \textbf{LLaVA-OneVision-7B} & \multirow{4}{*}{\textbf{CLIP}}
      & 60.6 & 59.6 & 59.2 & 59.8 & 59.0 & 59.3 & 59.5 & 59.3 & 44.1 & 44.9 & 43.8 & 44.3 & 67.8 & 68.8 & 68.9 & 68.5 \\
    \textbf{LLaVA-Video-7B}     &
      & 63.6 & 62.8 & 63.0 & 63.1 & 66.4 & 65.7 & 65.8 & 66.0 & 48.1 & 49.4 & 48.0 & 48.5 & 72.2 & 73.0 & 72.6 & 72.6 \\
    \textbf{Qwen2.5-VL-7B}      &
      & 64.0 & 64.4 & 64.8 & 64.4 & 65.4 & 65.4 & 65.8 & 65.5 & 47.6 & 46.0 & 47.3 & 47.0 & 72.4 & 71.1 & 71.8 & 71.8 \\
    \textbf{Qwen3-VL-8B}        &
      & 66.4 & 66.6 & 66.5 & 66.5 & 69.9 & 70.1 & 69.5 & 69.8 & 49.7 & 51.3 & 50.4 & 50.5 & 75.1 & 74.8 & 75.1 & 75.0 \\

    \midrule

    \textbf{LLaVA-OneVision-7B} & \multirow{4}{*}{\textbf{LongCLIP}}
      & 59.5 & 60.4 & 60.1 & 60.0 & 59.9 & 59.3 & 59.2 & 59.5 & 44.3 & 44.9 & 45.3 & 44.8 & 69.8 & 69.0 & 69.4 & 69.4 \\
    \textbf{LLaVA-Video-7B}     &
      & 63.7 & 63.1 & 62.7 & 63.2 & 65.4 & 65.5 & 65.1 & 65.3 & 49.3 & 49.1 & 48.9 & 49.1 & 74.4 & 74.0 & 74.2 & 74.2 \\         
    \textbf{Qwen2.5-VL-7B}      &
      & 66.4 & 65.3 & 64.7 & 65.5 & 65.3 & 66.1 & 66.2 & 65.9 & 47.5 & 47.1 & 46.6 & 47.1 & 73.3 & 73.1 & 72.9 & 73.1 \\
    \textbf{Qwen3-VL-8B}        &
      & 66.9 & 66.8 & 66.9 & 66.9 & 69.9 & 69.3 & 69.5 & 69.6 & 51.1 & 52.2 & 50.8 & 51.4 & 76.5 & 76.3 & 76.4 & 76.4 \\

    \bottomrule
  \end{tabular*}

  \vspace{-0.15cm}
\end{table*}
\begin{table*}[t]
\centering
\caption{Repeated-run results under different embedding models (CLIP/BLIP/LongCLIP) in Tab. \ref{tab:ablation_qwen_clip_blip}, reporting test1/2/3 and the mean to quantify run-to-run variability.}

\label{tab:prob_sampling_fluctuation_ranges}

\scriptsize
\setlength{\tabcolsep}{3.2pt}
\begin{tabular*}{\textwidth}{@{\extracolsep{\fill}} ll cccc cccc cccc @{}}
\toprule
\multirow{2}{*}{\textbf{Model}} 
& \multirow{2}{*}{\textbf{Dataset}}
& \multicolumn{4}{c}{\textbf{CLIP}}
& \multicolumn{4}{c}{\textbf{BLIP}}
& \multicolumn{4}{c}{\textbf{LongCLIP}} \\
\cmidrule(lr){3-6} \cmidrule(lr){7-10} \cmidrule(lr){11-14}
& 
& \textbf{test1} & \textbf{test2} & \textbf{test3} & \textbf{mean}
& \textbf{test1} & \textbf{test2} & \textbf{test3} & \textbf{mean}
& \textbf{test1} & \textbf{test2} & \textbf{test3} & \textbf{mean} \\
\midrule

\multirow{2}{*}{\textbf{Qwen3-VL-8B}}
& LongVideoBench 
& 66.4 & 66.6 & 66.5 & 66.5 
& 66.2 & 66.5 & 67.5 & 66.7
& 66.9 & 66.8 & 66.9 & 66.9 \\
& MLVU
& 75.1 & 74.8 & 75.1 & 75.0
& 75.0 & 75.6 & 75.6 & 75.4
& 76.5 & 76.3 & 76.4 & 76.4 \\
\midrule

\multirow{2}{*}{\textbf{LLaVA-Video-7B}}
& LongVideoBench
& 63.6 & 62.8 & 63.0 & 63.1
& 62.4 & 62.8 & 63.6 & 62.9
& 63.7 & 63.1 & 62.7 & 63.2 \\
& MLVU
& 72.2 & 73.0 & 72.6 & 72.6
& 73.5 & 73.3 & 73.3 & 73.4
& 74.4 & 74.0 & 74.2 & 74.2 \\

\bottomrule
\end{tabular*}

\vspace{-0.15cm}
\end{table*}

\subsection{Repeated-run Evaluation Results}
\label{app:specific_test}
In this appendix, we report detailed per-run results that are omitted in the main paper due to space limitations.
All numbers in Tab.~\ref{tab:main_results} are averaged over three repeated evaluations under identical inference settings (same backbone, frame budget, and evaluation protocol).
For transparency, Tab.~\ref{tab:flow} lists the score of each run together with the mean, allowing readers to directly inspect run-to-run variability induced by probabilistic sampling.
Furthermore, Tab.~\ref{tab:prob_sampling_fluctuation_ranges} provides the same three-run breakdown and mean results when varying the scoring backbones (CLIP/BLIP/LongCLIP), which complements the robustness trends summarized in Tab.~\ref{tab:ablation_qwen_clip_blip}.

\noindent\textbf{Per-run results for the main comparison.}
Tab.~\ref{tab:flow} presents the results of three repeated runs, along with their mean values across benchmarks, demonstrating that AdaQ consistently outperforms previous methods across multiple trials, despite the inherent randomness introduced by different random seeds.

\noindent\textbf{Repeated-run results under different scoring backbones.}
Tab.~\ref{tab:prob_sampling_fluctuation_ranges} further evaluates stability when changing the embedding model used for frame--query relevance (CLIP/BLIP/LongCLIP), and shows that the observed improvements remain reliable across scoring backbones, consistent with Tab.~\ref{tab:ablation_qwen_clip_blip}.

\section{Compute Resources and Runtime Analysis}
\label{app:efficient}
\begin{table}[!t]
\centering
\caption{
Efficiency comparison between \emph{Uniform} sampling and \emph{AdaQ} on \emph{Qwen3-VL-8B}.
\emph{Candidate Frames} denotes the frames used for CLIP-based similarity computation, and \emph{Input Frames} denotes the final frames fed into the MLLM.
}
\label{tab:efficiency_results}

\resizebox{\textwidth}{!}{%
\scriptsize
\setlength{\tabcolsep}{2.6pt}
\renewcommand{\arraystretch}{0.95}
\begin{tabular*}{\textwidth}{@{\extracolsep{\fill}} l c c c c c c c @{}}
\toprule
\textbf{Method}
& \textbf{Candidate Frames}
& \textbf{Input Frames}
& \textbf{Memory Overhead}
& \textbf{CLIP Cost}
& \textbf{Selection Cost}
& \textbf{MLLM Inference}
& \textbf{All Time}
\\
\midrule

\textit{Uniform}
& -
& 64
& 17.98 GB
& -
& -
& 2.61s
& 21.3s
\\

\textit{Uniform}
& -
& 256
& 22.83 GB
& -
& -
& 7.80s
& 26.8s
\\

\midrule

\textit{AdaQ}
& 512
& 64
& 17.98 GB
& 0.53s
& 0.001s
& 2.67s
& 21.89s
\\

\textit{AdaQ}
& 1024
& 64
& 17.98 GB
& 0.66s
& 0.001s
& 2.61s
& 22.04s
\\

\bottomrule
\end{tabular*}%
}
\vspace{-0.2cm}
\end{table}
Although AdaQ requires computing frame-query similarities over all candidate frames, this cost is also shared by most existing query-aware frame selection methods. Moreover, this step is performed by CLIP-like VL embedding models, which are much smaller and computationally cheaper than MLLMs. In practice, the runtime is still dominated by MLLM inference rather than similarity scoring. Therefore, the additional cost of candidate-frame scoring is relatively minor compared with the cost of directly feeding more video frames into the MLLM. To provide a concrete runtime analysis, we report the actual cost of AdaQ on Qwen3-VL-8B using one NVIDIA A800 GPU in Table~\ref{tab:efficiency_results}. Compared with uniform sampling using 256 input frames, AdaQ only feeds 64 selected frames into the MLLM and thus keeps the memory overhead at 17.98 GB, while uniform sampling with 256 frames increases the memory overhead to 22.83 GB. Moreover, the selection step of AdaQ is nearly negligible, taking only 0.001s, and the total runtime remains close to uniform sampling with 64 frames even when 512 or 1024 candidate frames are scored. These results show that AdaQ introduces little selection overhead while avoiding the much higher inference cost caused by directly increasing the number of MLLM input frames.

\section{Limitations}
\label{app:limitations}
Although AdaQ achieves consistent improvements, its performance is still limited by two factors. First, some failures arise from the capability boundary of the underlying MLLM, especially for complex temporal reasoning or ambiguous visual content. Second, AdaQ relies on external VL embedding models for frame-query similarity estimation, where adaptive sampling can alleviate but not fully resolve noisy or misaligned similarities. Future work may combine adaptive frame sampling with stronger temporal reasoning and video-language alignment.

\section{Broader Impacts}
\label{app:impact}
AdaQ aims to improve the efficiency and effectiveness of long-video understanding for MLLMs by reducing redundant visual inputs and enabling more adaptive frame sampling. This may lower the computational cost of video-based AI systems and make long-video understanding more accessible in resource-limited scenarios. However, stronger and more efficient video understanding models may also introduce potential risks when used in privacy-sensitive applications, such as surveillance or large-scale video analysis without proper consent. In addition, AdaQ relies on pretrained MLLMs and VL embedding models, and may inherit their biases, hallucinations, or failure modes, which could lead to unreliable predictions in high-stakes scenarios. Therefore, real-world deployment should follow dataset licenses, privacy regulations, and appropriate human oversight.

\section{Licenses for Existing Assets}
\label{app:license}
This work builds upon existing public datasets, pretrained models, VL embedding models, and baseline codebases. Specifically, we use public long-video understanding benchmarks, including LongVideoBench, Video-MME, LVBench, and MLVU, as well as publicly available MLLM backbones and VL embedding models, such as LLaVA-OneVision, LLaVA-Video, Qwen2.5-VL, Qwen3-VL, CLIP, LongCLIP, and BLIP. We properly cite the original papers and official releases of these assets in the main paper. All datasets, models, and codebases are used only for research purposes, and we follow their official licenses, terms of use, and evaluation protocols. Our released code does not redistribute the original datasets or pretrained model weights, but instead provides instructions and scripts for using the assets from their official sources. We will also include the corresponding asset names, sources, and license or usage information in the released project documentation.





\newpage

\end{document}